\documentclass[conference]{IEEEtran}
\IEEEoverridecommandlockouts
\usepackage{cite}
\usepackage{amsmath,amssymb,amsfonts}
\usepackage{algorithmic}
\usepackage{graphicx}
\usepackage{textcomp}
\usepackage{xcolor}
\usepackage{url}
\usepackage{booktabs}
\usepackage{hyperref} 
\usepackage{adjustbox}
\usepackage{graphicx}
\usepackage{pythonhighlight}
\usepackage{listings}

\usepackage{fancyhdr}
\begin{document}

\title{GSCE: A Prompt Framework with Enhanced Reasoning for Reliable LLM-driven Drone Control
}

\author{\IEEEauthorblockN{Wenhao Wang}
\IEEEauthorblockA{University of Massachusetts Dartmouth\\
wwang5@umassd.edu}
\and
\IEEEauthorblockN{Yanyan Li}
\IEEEauthorblockA{California State University San Marcos\\
yali@csusm.edu}
\and
\IEEEauthorblockN{Long Jiao and Jiawei Yuan}
\IEEEauthorblockA{University of Massachusetts Dartmouth\\
\{ljiao, jyuan\}@umassd.edu}
}

\maketitle \thispagestyle{fancy}


\begin{abstract}
The integration of Large Language Models (LLMs) into robotic control, including drones, has the potential to revolutionize autonomous systems. Research studies have demonstrated that LLMs can be leveraged to support robotic operations. However, when facing tasks with complex reasoning, concerns and challenges are raised about the reliability of solutions produced by LLMs. In this paper, we propose a prompt framework with enhanced reasoning to enable reliable LLM-driven control for drones. Our framework consists of novel technical components designed using \underline{G}uidelines, \underline{S}kill APIs, \underline{C}onstraints, and \underline{E}xamples, namely \emph{GSCE}. GSCE is featured by its reliable and constraint-compliant code generation. We performed thorough experiments using GSCE for the control of drones with a wide level of task complexities. Our experiment results demonstrate that GSCE can significantly improve task success rates and completeness compared to baseline approaches, highlighting its potential for reliable LLM-driven autonomous drone systems.
\end{abstract}



\section{Introduction} \label{Intro}
The rapid advancement of LLM has led to the revolution of a wide range of embodied agents. Models such as GPT4 \cite{GPT4}, Llama 3 \cite{Llamma}, and Gemini \cite{Gemini} have achieved remarkable results in various robotic tasks, such as planning \cite{planning}, navigation \cite{Navi, L3MVN}, communication \cite{Survey}, and automation \cite{automation}. Given the recent advances of LLMs and their remarkable capabilities of natural language (NL) understanding and context generation, how to leverage LLMs for drones has become an emerging topic to bring more intelligence to robotic systems. Recently, research efforts have been made towards the reliable integration of LLMs and drones \cite{ChatGPTRobotics, TypeFly, CLEAR, AdaptiveRAG}, promising to revolutionize LLM-driven autonomous drone control.

Although LLMs have demonstrated potential for integration into drone control, they still face challenges in logical reasoning \cite{logicalreasoning}, which limits their reliability in operations requiring complex decision-making. Unlike many other text generation applications where semantic-level accuracy is sufficient, drone control necessitates the precise execution of a sequence of actions. If the control code generated by LLMs is incorrect or invalid, it may lead to erroneous drone behavior, posing safety and security risks to the public, such as drone crashes. These limitations also hinder their adaptability to dynamic environments, where adherence to human instructions is essential.

\begin{figure}[t]
    \centering
    \includegraphics[width=.9\linewidth]{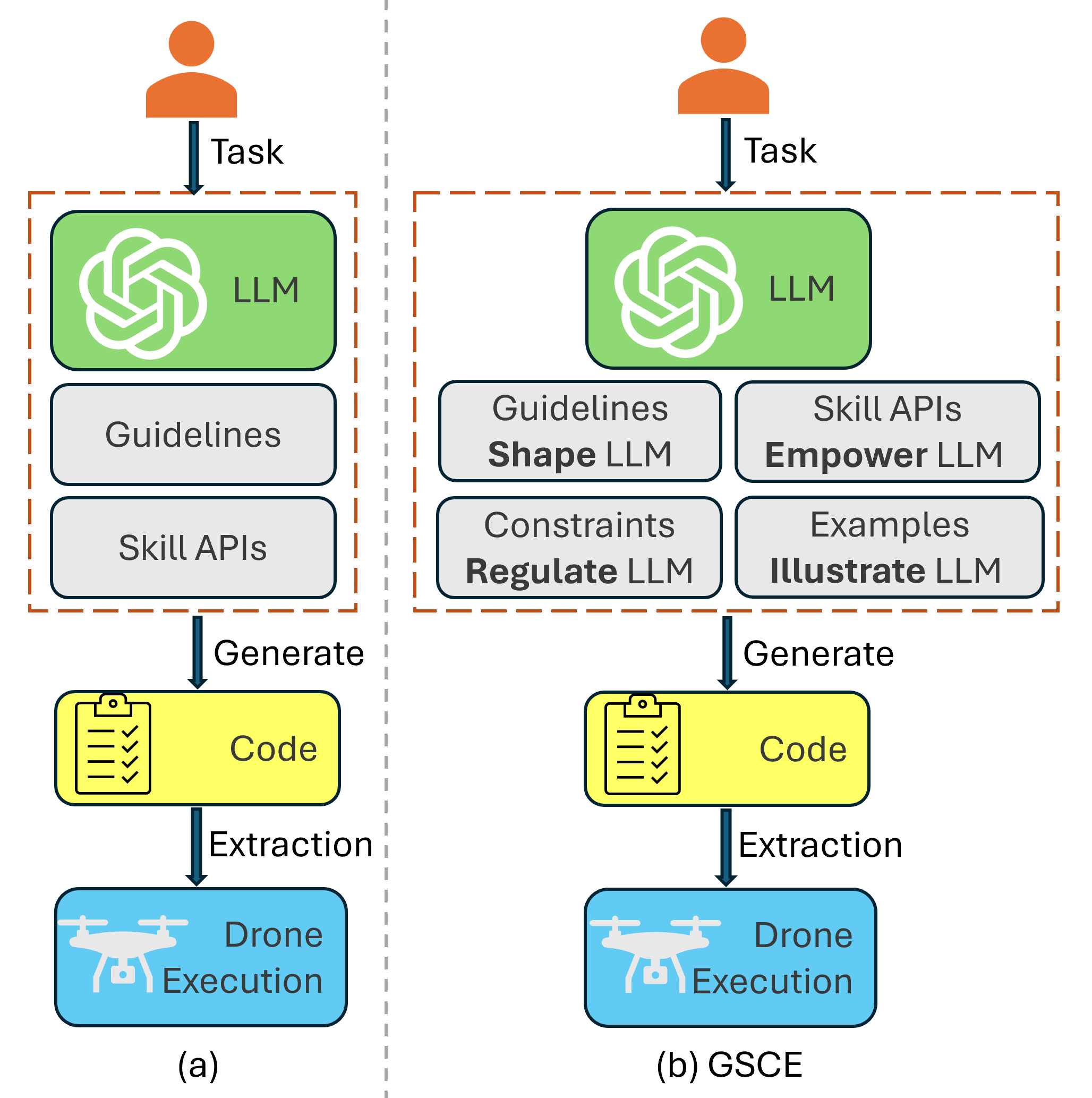}
    \caption{Illustration of LLM-driven drone control systems prompt structures. Structure (a) prompts LLMs with guidelines and skill APIs, as demonstrated in \cite{ChatGPTRobotics, CodeasPolicies}. Our proposed GSCE structure (b) introduces constraints to regulate LLM behavior and incorporates examples with constraints implementation to illustrate LLM, thereby enhancing the reasoning capabilities of LLMs.}
    \label{fig:Overview}
\end{figure}

A prevalent approach to overcome this issue involves prompting LLMs with guidelines and skill APIs, leveraging LLMs' contextual understanding to interpret and apply these resources effectively. By adhering to the provided guidelines and utilizing the appropriate APIs, LLMs can generate code that controls drones to operate in accordance with human instructions, as shown in Fig.\ref{fig:Overview}(a). However, this approach is not reliable for complex reasoning tasks and poses risks such as incorporating incorrect APIs or misusing them in the generated code. Moreover, LLMs struggle to reason the correct actions in scenarios where LLMs have limited prior knowledge or must operate within robotic constraints.

To address these challenges, we propose a novel prompt framework designed to enhance the reasoning capabilities of LLM-driven drone control. Our approach incorporates a structured prompt framework consisting of Guidelines, Skill APIs, Constraints, and Examples. As illustrated in Fig. \ref{fig:Overview}(b), the \textbf{Guidelines} shape the LLM’s characteristics for code generation in drone control, while the \textbf{Skills APIs} are predefined function interfaces that empower LLM with essential drone control capabilities. The \textbf{Constraints} regulate actions in scenarios where the LLM lacks prior knowledge or where the Skill APIs do not document specific requirements. Lastly, the \textbf{Examples} leverage the in-context learning \cite{URIAL} capability of LLMs to illustrate how task descriptions should be mapped to control code. Additionally, the examples demonstrate the implementation of NL constraints to ensure compliance with predefined constraints while reasoning. Our contribution can be summarized as follows:

\begin{itemize}
    \item We propose a novel prompt framework designed to enhance the reasoning capabilities of LLM-driven drone control.
    \item Beyond using illustrative examples, our approach explicitly incorporates constraints within examples to further enhance LLM reasoning.
    \item The framework is modular, with each component dynamically adaptable, allowing for seamless integration across diverse robotic platforms and task scenarios.
\end{itemize}

\section{Related Work} \label{RelatedWork}

\subsection{LLM for Robotics}
LLMs have demonstrated remarkable potential in semantic understanding and context generation, making them a valuable asset in robotics. ChatGPT for Robotics \cite{ChatGPTRobotics} and Code as Policies \cite{CodeasPolicies} achieve remarkable progress in using LLM-generated code to control robotic arms. For drone applications, TypeFly \cite{TypeFly} proposes an end-to-end system for piloting a quadcopter with LLM, while REAL \cite{REAL} provides a strategy to employ LLM as part of control and planning on drones. Additionally, LLM can act as a cognitive agent in closed-loop vehicle motion planning. \cite{PlanAgent} These efforts highlight the diverse and impactful applications of LLMs across different robotic platforms. However, robotic performance remains constrained by the reasoning capabilities of LLMs. LLMs often exhibit unreliability in tasks requiring complex reasoning.  Notably, ChatGPT has acknowledged the necessity of human intervention for code generation \cite{ChatGPTRobotics}. 

\subsection{Prompt Engineering}
Recent research has increasingly focused on leveraging prompt engineering \cite{PromptEngineering} strategies to enhance the reasoning capabilities of LLMs for robotics. In ChatGPT for Robotics \cite{ChatGPTRobotics}, LLMs are provided with a description of guidelines, constraints, and accessible APIs, and then LLMs generate code to control robots completing the given task description. Similarly, ViLaIn \cite{VisionPlanning} provides detailed descriptions of the robot and its environmental states, helping LLMs develop a general understanding of the robot's condition before generating control code. However, LLMs exhibit sensitivity to input perturbations \cite{JailbreakingLLM, PromptInjection, HighlightSafetyConcern}. To address this, PluginSafety \cite{PluginSafety} applies safety chips that infer specification constraints, ensuring that LLM-generated content adheres to predefined NL constraints. AutoTamp \cite{AutoTAMP} adopts a different approach by prompting the LLM to first generate an intermediate task representation and then translate it into task plans. Additionally, some studies integrate human or automated checker feedback \cite{AutoTAMP, ISR-LLM, Lifelong, EnsureSafety} to re-prompt the LLM, refining its outputs for improved reasoning accuracy.

\subsection{LLM In-context Learning}
In-context learning is a powerful capability of LLMs that enables them to perform tasks by leveraging examples provided in the input prompt. Instead of retraining model parameters, LLMs recognize patterns from the given context and generalize them to new outputs, allowing users to shape the characteristics of LLMs with few examples \cite{URIAL}. In robotics, in-context learning enables LLMs to generate code following robot policy by incorporating example demonstrations within the prompt  \cite{CodeasPolicies}. This paradigm approach helps LLMs generalize across tasks by analyzing prior examples. Furthermore, LLMs can learn to ground human instructions using the provided APIs from few-shot examples \cite{PromptBook}. However, most existing studies primarily focus on constructing examples to illustrate LLM, often overlooking their integration with prompt engineering to further enhance performance.

\subsection{Chain of Thought (COT) Reasoning}
CoT \cite{CoT} encourages LLMs to articulate their intermediate reasoning step by step when solving a task. Given that robotic tasks often require executing a series of sequential actions, CoT facilitates the generation of code that aligns with each step of the action plan. Prior research incorporates CoT into the comment of code examples \cite{PromptBook}, while another study utilizes CoT for translating plans into actionable steps \cite{ISR-LLM}. 

\section{Problem Description}
 In this research, our primary focus is on the reasoning capability of LLMs. Following a prevalent approach, the LLM generates code directly without first mapping intermediate plans to executable code. 
 
 Given a task description (query), the LLM is required to reason through the task and generate code that follows the specified NL movement instructions. The reasoning problem is formulated as: \(P = \langle init\_s, S, A, T, G \rangle\)
, where the problem \(P\) consists of discrete states, state transition, and a finite set of actions. For a given state \( S \), an action \( a \) is selected from the skill set \( A \), leading to state transition: \(T: S \times A \rightarrow S\) that determines the next state. Here, \( init\_s \) represents the initial state, and \( G \subseteq S\) denotes the goal states. A solution to this problem is a sequence of actions \(l = (a_1, a_2, a_3, \dots, a_n)\) that control the drone to fly from the initial state through intermediate states until reaching the goal state. The state consists of the drone’s position and attitude, while the actions correspond to drone maneuvers.

\section{Design of GSCE} \label{Method}
\begin{figure*}[ht]
    \centering{\includegraphics[scale=0.19]{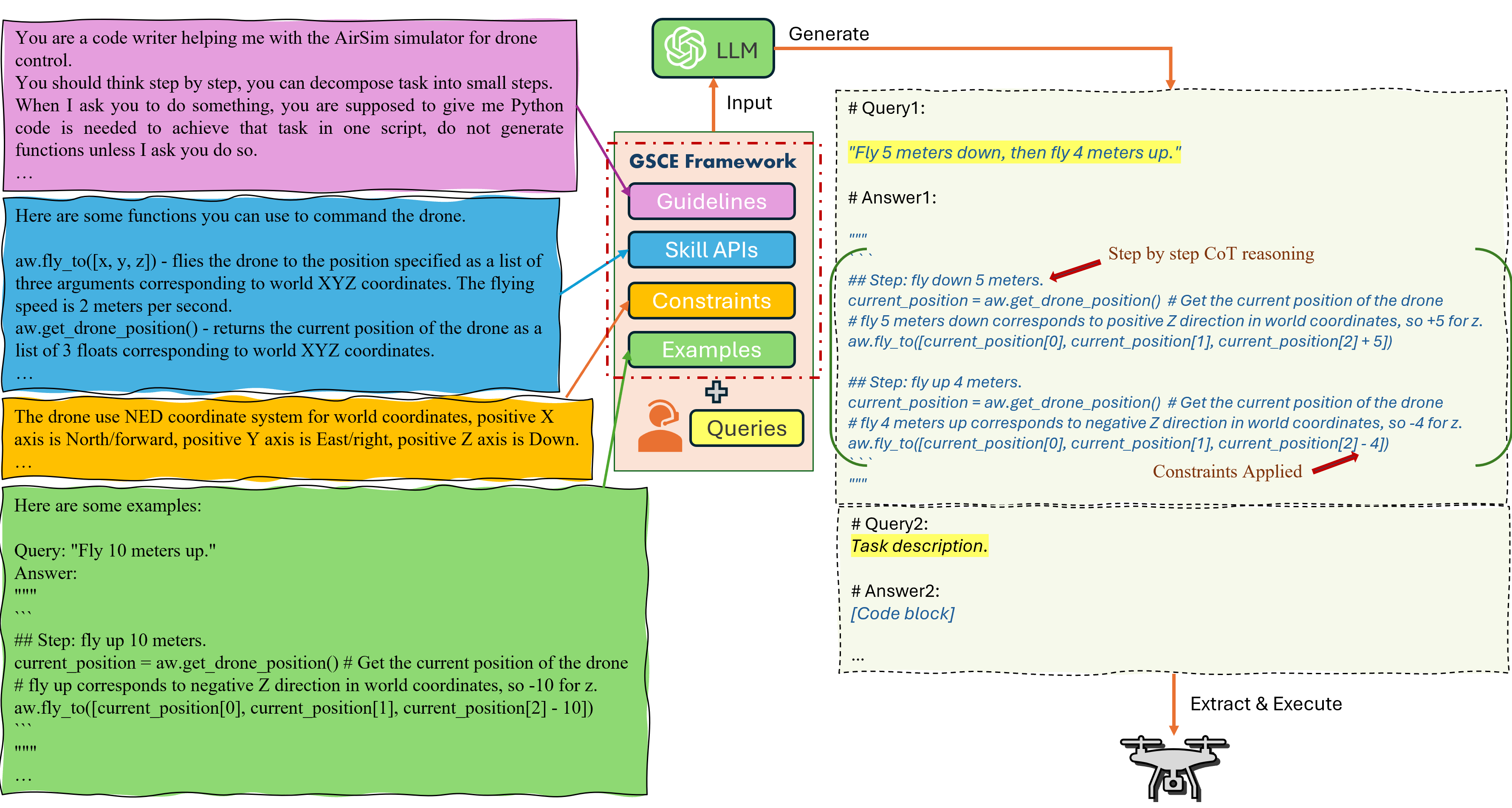}}
    \caption{Our method starts by establishing the GSCE framework with guidelines, skill APIs, constraints, and examples. The user then provides queries (task descriptions) to the LLM for code generation. The generated code is extracted from the output context and executed to control the drone. Within the generated code, the LLM generates a sequence of actions that employ step-by-step reasoning and implement the constraints correctly.}
    \label{fig:Method}
\end{figure*}

Our key insight is to develop a robust method that enhances reasoning capabilities for LLM in drone operations. In contrast to previous work that only prompted limited context to LLM \cite{TypeFly, ChatGPTRobotics, REAL}, our approach explicitly incorporates the usage of constraints within the examples to enhance LLM reasoning capabilities. This method enables the LLM to develop a comprehensive understanding of the constraints and their application, thereby enhancing the LLM's ability to reason and apply these constraints in new tasks appropriately. To achieve this, we introduce a structured prompt framework, referred to as GSCE, which consists of guidelines, robotic skill APIs, constraints, and examples. When given an NL task description, the LLM is first prompted with the GSCE framework before generating executable code. The following subsections provide a detailed explanation of the GSCE Prompt Framework.

\subsection{GSCE Prompt Framework Overview}
As presented in Fig.\ref{fig:Method}, the GSCE prompt framework is composed of four key components: guidelines, skill APIs, constraints, and examples. An advantage of this framework is its support for the dynamic adaptation of each component, allowing seamless integration across diverse robotic platforms and task scenarios. Below is an overview of each component:
\begin{itemize}
    \item \textbf{Guidelines} outline the role of the LLM as an assistant for drone control code generation and establish the desired coding style, ensuring consistency and clarity in the generated code. 
    \item \textbf{Robot skills APIs} provide detailed instructions on the available APIs that the LLM can utilize when generating code. These instructions include the names of APIs and explanations of their functionalities, allowing the LLM to employ the appropriate functions for a given task.
    \item \textbf{Constraints} encapsulate domain-specific policies that extend beyond the APIs. They address scenarios where the LLM's prior knowledge is insufficient or where adherence to specific implementation standards is required.
    \item \textbf{Examples} serve as implicit demonstrations, showing the LLM how NL task queries and constraints can be translated into code solutions using the provided APIs. Additionally, CoT is embedded within the examples, guiding the LLM to reason through tasks step by step.
\end{itemize}

\subsection{Shaping LLM Characteristics with Guidelines} \label{Guidelines}
To leverage the generalization capabilities of LLMs, guidelines provide foundational instructions outlining how the LLM should behave and respond in generalization \cite{SystemPrompt}. In the GSCE prompt framework, guidelines initially establish the role of the LLM as an assistant specifically for robot control code generation. They then extend to specify robot embodiments (drones) and task information. To mitigate the risk of hallucination, the guidelines require the LLM to utilize assigned functions rather than hypothesized or self-generated functions. Additionally, the guidelines emphasize adherence to a standardized code format to ensure consistency and regularity. To enhance robustness, the guidelines include provisions to prevent the occurrence of exceptions or infinite loops in generated code. As indicated in the purple block of Fig.\ref{fig:Method}, these guidelines outline the LLM’s role as a code generator for drone control while preventing it from hypothesizing new functions.

\subsection{Empower LLM with Skill APIs} \label{Skills}
After establishing the guidelines for drone control code generation, the LLM is prompted with a set of Skill APIs that it can leverage during code generation. To align with prevalent LLM for robotics studies that required LLM to utilize designated APIs for generating executable code\cite{ChatGPTRobotics, AdaptiveRAG}, the GSCE prompt framework introduces a collection of skill APIs specifically designed for drone control. These APIs encapsulate essential drone maneuvering skills, such as takeoff, landing, navigation to specific coordinates, yaw angle adjustment, and retrieving state information from the drone. Furthermore, the skill APIs not only list the available skills but also provide detailed documentation on each API’s usage and functionalities, ensuring the LLM develops a comprehensive understanding of the skills. Prompt LLM with skill APIs empower the LLM to effectively select and apply appropriate APIs during code generation. As illustrated in the blue block of  Fig.\ref{fig:Method}, the Skill APIs component empowers LLM with a predefined action set \(A = (fly\_to(), get\_drone\_position(), ...)\) that can be utilized to generate code for drone control. The right section of Fig.\ref{fig:Method} further demonstrates how the LLM accurately selects and applies these actions to ensure the state transitions \(T\) adhere to the specified task objectives.

\subsection{Regulate LLM with Constraints} \label{Constraints}
LLM may fail to adhere to human instruction when provided solely with guidelines and Skill APIs \cite{ChatGPTRobotics}. Consequently, constraints play a crucial role in regulating robot policies, particularly in scenarios where the explanation of robot skills APIs is insufficient or the LLM lacks prior knowledge. These constraints encapsulate domain-specific rules or robot requirements that are essential for ensuring LLM reasons accurately, reducing the risk of generating unsafe control code. In our approach, the LLM is provided with constraints on the coordination directional information that the LLM has limited prior knowledge. Additionally, considering the inherent challenges LLMs face in spatial reasoning \cite{SpatialReasoning}, these constraints can serve as explicit instructions to support the LLM for reasoning tasks. Highlighted in the orange block of Fig.\ref{fig:Method}, constraints provide directional information that regulates the LLM in spatial direction reasoning. However, constraints alone can be vulnerable to ``jailbreaking" \cite{JailbreakingLLM} or forgotten \cite{CatastrophicForgetting} while generating code. To address these issues, we further augment constraints with illustrative examples in the GSCE prompt framework, which is elaborated in the next subsection \ref{Enforce Constraints}.

\subsection{Illustrate LLM via Examples} \label{Examples}
Examples are the key component in LLM for robotics \cite{PromptBook, CodeasPolicies}, they leverage the in-context learning capabilities of LLMs. In addition to demonstrating LLM as prior studies, examples in GSCE extend to enhance the LLM's understanding of NL constraints and their implementations in code generation. They also enable the LLM to effectively learn and replicate the desired coding patterns from example code solutions. Consequently, the reasoning capabilities of the LLM are improved through imitating well-crafted examples. 

In the GCSE prompt framework, there are \(K\) examples. Each example consists of an NL task description (query) and a corresponding Python code solution (answer) that utilizes the robot skills APIs in section \ref{Constraints}. The design of examples is guided by two key factors:

\subsubsection{\textbf{Enforcing Constraints by Examples}} \label{Enforce Constraints}
Contextual constraints are frequently overlooked in code generation \cite{FLTRNN}. To address this limitation, we build on the principles of emphasizing the critical role of safety examples in aligning LLM with safety standards \cite{PTST}. Within the GSCE prompt framework, constraints are integrated not only as contextual policies but also as implementations within code solutions. The provided examples comprehensively cover all constraints specified in the NL constraints section. Additionally, code solutions are annotated with comments that specify the applied constraints for LLM to comprehend. By prompting the LLM with well-structured examples, the framework enforces LLM adherence to constraints through in-context learning, ensuring that the generated code aligns with constraints. In the green block of Fig.\ref{fig:Method}, an example implementation of directional constraints (as defined in the orange block) enables the LLM to learn how directional constraints are implemented and subsequently ensure the drone reaches the goal state \(G\) in the right section.

\subsubsection{\textbf{CoT Reasoning}}
CoT reasoning is a popular prompting strategy that enables LLM to perform reasoning tasks in a step-by-step approach. Given the nature of robotic tasks that are sequentially executed actions, integrating CoT prompting enhances LLM reasoning ability by decomposing a robotic task into a chain of steps. In the GSCE examples, the task is decomposed into sequential movement steps, with the code solutions structured to align with each step. Additionally, the reasoning progress of each step of movement is documented as comments elucidating how each step of thought is grounded. This approach enables LLM to reason step-by-step as demonstrated in examples and apply CoT reasoning when given a new task. As shown in Fig.\ref{fig:Method}, LLM imitates illustrations from examples (green block), which effectively decompose new tasks into discrete steps that correspond to a sequence of actions \(l = (a_1 = \text{fly down 5 meters}, a_2 = \text{fly up 4 meters})\).

\section{Experiments}
The objective of this experiment is to evaluate the hypothesis that our proposed approach enhances the reasoning capabilities of LLMs. To achieve this, the experiment is designed with three primary goals: \((1)\) to demonstrate the implementation of an LLM-based robot system in a simulated environment, \((2)\) to assess the reasoning abilities of our designed LLM-based robot system in the context of drone manipulation tasks, and \((3)\) to compare the system's performance under various configurations of components.

\subsection{Experiment Setup}
\subsubsection{\textbf{Experimental Environment}} The experiment is conducted on a quadcopter using ``simple\_flight" flight controller \cite{SimpleFlight} in a simulated environment AirSim \cite{AirSim}. AirSim provides realistic environments for testing and developing autonomous systems such as drones and self-driving cars, making it well-suited for this research. All tasks were experimented with in the pre-built environment named ``Block"\cite{Blocks}. It has a large flying area with several cubic structures, providing users with an accessible platform for basic testing and execution of drone control scripts. During the experiment, drone state information was accessible from the simulator and utilized for evaluation purposes.

\subsubsection{\textbf{Tasks}}
Unlike prior studies that primarily focused on simple drone movements (e.g., fly to a position), the proposed tasks comprehensively evaluate the reasoning capabilities of the LLM by encompassing a wide level of complexities.  Each task necessitates step-by-step reasoning to analyze the spatial relationships between the target and initial drone states. 

A total of 44 NL task descriptions were designed, requiring the LLM to generate code for controlling drone movements in AirSim according to the task objectives. These tasks involve both basic movements (e.g., straight-forward) and complex ones that require adherence to one or more robotic policy constraints. Three sample tasks used in our experiment with different complexities are provided below:
\begin{itemize}
    \item \textit{``Fly 5 meters down, then fly 4 meters up."}
    \item \textit{``Turn 90 degrees clockwise, then fly 4 meters left in the drone's body frame."}
    \item \textit{``Fly the drone in the top-right direction at an angle of 30 degrees from the horizontal axis, in the YZ plane of drone's body frame for a distance of 10 meters."} 
\end{itemize}
Additionally, each task was manually validated in AirSim to ensure feasibility, avoiding potential errors inherent to the AirSim simulator that could affect the evaluation process.

\subsubsection{\textbf{GSCE Prompt Framework Implementation}}
First, the LLM was guided for drone control code generation in AirSim using Python. The model was then provided with a set of fundamental skill APIs that it could leverage during code generation, including \(takeoff()\), \(land()\), \(get\_yaw()\), \(set\_yaw()\), \(fly\_to()\), and \(get\_drone\_position()\).  Subsequently, the LLM was introduced to detailed NL constraints related to drone/world directional information and the policies for coordinate transformation. Finally, three \((K=3)\) pairs of task descriptions and code solutions with CoT were provided as examples.

\subsubsection{\textbf{Compared Methods}}
We compare four methods for drone control in the AirSim simulator. The configurations associated with each method are summarized in Table \ref{tab:ComparedMethod}:

\begin{table}[h]
    \centering
    \caption{Configuration of Compared Methods}
    \begin{adjustbox}{width=0.48\textwidth}
    \begin{tabular}{l|c|c|c|c}
        \toprule
         & Base \cite{ChatGPTRobotics} & Constraints \cite{DoasICan} & Examples \cite{CodeasPolicies} & GSCE \\
        \midrule
        Guidelines & \checkmark & \checkmark & \checkmark & \checkmark \\
        Skill APIs & \checkmark & \checkmark & \checkmark & \checkmark \\
        NL Constraints &  & \checkmark &  & \checkmark\\
        Examples &  &  & \checkmark & \checkmark \\
        \bottomrule
    \end{tabular}
    \end{adjustbox}
    \label{tab:ComparedMethod}
\end{table}

\begin{itemize}
    \item \textbf{LLM-Direct (Base):} This baseline approach is adapted from ChatGPT for Robotics \cite{ChatGPTRobotics} and serves as the foundation for the subsequent methods. In LLM-Direct, the LLM is prompted with general guidelines for generating AirSim drone control code.
    \item \textbf{LLM-NL\_constraints (Constraints) :} This method extends LLM-direct by incorporating some NL constraints to regulate the LLM in generating code that adheres to constraints \cite{DoasICan}.
    \item \textbf{LLM-Examples (Examples):} Inspired by \cite{CodeasPolicies}, this approach builds on LLM-Direct by providing illustrative examples \((K=3)\) that demonstrate constraints within the context of code solutions. Unlike LLM-NL\_Constraints, this method does not include explicit NL constraints.
    \item \textbf{GSCE:} The LLM is provided with both NL constraints and examples along with their corresponding solutions, as outlined in Section. \ref{Examples}.
\end{itemize}

For all the aforementioned methods, \textit{``gpt-4-turbo-2024-04-09"} (GPT-4-Turbo) and \textit{``gpt-4o-2024-08-06"} (GPT4o) serve as the backbone LLMs. All methods employ an end-to-end code generation approach; in other words, each task description was provided as input to the LLM, which then generated the corresponding drone control code. The generated code was subsequently executed in AirSim to evaluate the reasoning performance of each method. To ensure consistency in the evaluation, code generation for each task was repeated three times.

\subsection{Evaluation Setup}
\subsubsection{\textbf{Metrics}}
The performance of the abovementioned methods is evaluated using two key metrics: \textbf{Success Rate (SR)} and \textbf{Completeness}. Unlike prior studies \cite{FLTRNN, PluginSafety}, which primarily assess whether the task's final goal is reached, our \textbf{Success Rate} considers all goal states, including intermediate states and the final state. In the real-world scenario, the drone should fly the trajectory precisely as human instructed. Therefore, a task is counted as successful only when the drone reaches the final state while following the correct sequence of intermediate state transitions of the ground truth trajectory. The \textbf{Completeness} quantifies the percentage of goal states successfully reached, calculated as the ratio of correct state transitions to the total number of required state transitions of the ground truth trajectory in each task. Completeness examines the percentage of intermediate state transitions the LLM reasoned correctly, offering insights into its reasoning capability across all goal states.

\subsubsection{\textbf{Ground Truth}}
The ground truth for each task is represented by a list of state transitions. Each state transition is defined by four parameters: \([X, Y, Z, Yaw]\). Here, \(X\), \(Y\), and \(Z\) denote the positional changes of the drone along the world coordinate axes, while \(Yaw\) represents the change in the drone's yaw angle. These parameters capture the magnitude of change resulting from each movement.

\begin{table}[b]
    \centering
    \caption{Average Success Rate and Completeness of Proposed Methods}
    \begin{tabular}{lcccc}
        \toprule
         & \multicolumn{2}{c}{GPT-4-Turbo} & \multicolumn{2}{c}{GPT4o} \\
        \midrule
         & SR & Completeness & SR & Completeness \\
        \midrule
        Base \cite{ChatGPTRobotics} & 7.6\% &31.1\% & 17.4\% & 31.4\%\\
        Constraints \cite{DoasICan} & 51.5\% & 64.0\% & 59.8\% & 70.4\%\\
        Examples \cite{CodeasPolicies} & 66.7\% & 77.7\% & 65.9\% & 73.9\%\\
        GSCE & \textbf{90.9}\% & \textbf{92.0}\% & 89.4\% & 91.3\%\\
        \bottomrule
        
    \end{tabular}

    \label{tab:table1}
\end{table}

\subsection{Overall Results}

The experiment results are presented in Table \ref{tab:table1}, demonstrating a substantial improvement in both success rate and completeness when NL constraints and illustrative examples are incorporated into the base model. Notably, the GSCE framework outperforms all other methods, achieving the highest performance. In contrast, the base model is inadequate for reasoning tasks, highlighting its limitations in reasoning complex tasks.

Furthermore, the Constraints and Examples methods significantly improve performance, indicating that the inclusion of either contextual constraints or illustrative examples enhances the LLM’s performance by leveraging its in-context learning capabilities. However, the success rate remains below 70\%, and completeness does not exceed 80\%. In contrast, the GSCE framework significantly improves the reasoning capabilities of the LLM by integrating constraints with illustrative examples, achieving a success rate and completeness above 90\% on GPT-4-Turbo. These results highlight the effectiveness of combining NL constraints with illustrative examples to maximize the reasoning capabilities of the LLMs.

Additionally, we observed that GPT-4-Turbo performs slightly better than GPT-4o in configurations involving examples, suggesting that GPT-4-Turbo possesses better in-context learning capability. Despite these differences, both models show similar trends in performance improvement, indicating that the GSCE approach is generalizable across different LLMs.

\subsection{Design of Examples in GSCE}
This evaluation examines the design of examples within the GSCE framework. It is conducted with 
\(K=3\) examples, while all other experimental configurations remain unchanged.

\subsubsection{\textbf{Influence of CoT in Examples}} The results summarized in Table \ref{tab:cot} demonstrate that incorporating CoT within examples offers significant improvements in both success rate and completeness. This indicates that integrating CoT enables the LLM to perform step-by-step problem decomposition and reasoning through learning from examples. Additionally, CoT reasoning reduces the performance gap between GPT-4-Turbo and GPT-4o, suggesting its effectiveness in improving reasoning capabilities across different LLMs.

\begin{table}[h]
    \centering
    \caption{Average Success Rate and Completeness over the Influence of CoT in Examples}
    \label{tab:cot}
    \begin{adjustbox}{width=0.48\textwidth}
        \begin{tabular}{lccccc}
            \toprule
             & & \multicolumn{2}{c}{GPT-4-Turbo} & \multicolumn{2}{c}{GPT4o} \\
            \midrule
             & CoT & SR & Completeness & SR & Completeness \\
            \midrule
            Examples \cite{CodeasPolicies} & & 66.7\% & 77.7\% & 65.9\% & 73.9\%\\
            Examples \cite{CodeasPolicies} & \checkmark & \textbf{72.0}\% & \textbf{79.2}\% & 78.0\% & 79.5\%\\
            \midrule
            GSCE & & 81.1\% & 86.7\% & 78.0\% & 80.3\%\\
            GSCE & \checkmark & \textbf{90.9}\% & \textbf{92.0}\% & 89.4\% & 91.3\%\\
        \bottomrule
        \end{tabular}
    \end{adjustbox}
    
\end{table}

\subsubsection{\textbf{Influence of Constraints Implementation in Examples}} Results in Table \ref{tab:constraints} underscore the importance of integrating constraints within examples to optimize LLM reasoning performance, demonstrating that the LLM effectively applies constraints in the generated code.

\begin{table}[h]
    \centering
    \caption{Average Success Rate and Completeness over the Influence of Constraint Implementation in Examples}
    \begin{adjustbox}{width=0.475\textwidth}
    \begin{tabular}{lccccc}
        \toprule
         & & \multicolumn{2}{c}{GPT-4-Turbo} & \multicolumn{2}{c}{GPT4o} \\
        \midrule
         & Constraints & SR & Completeness & SR & Completeness \\
        \midrule
        GSCE & & 53.8\% & 71.6\% & 61.4\% & 74.2\%\\
        GSCE & \checkmark & \textbf{90.9}\% & \textbf{92.0}\% & 89.4\% & 91.3\%\\
    \bottomrule
    \end{tabular}
    \end{adjustbox}
    \label{tab:constraints}
\end{table}

Notably, the GSCE framework consistently outperforms the Examples method, regardless of whether CoT is applied. This result highlights the complementary effect of integrating NL constraints with illustrative examples, which together improve the reasoning capabilities of LLMs.

\subsection{Number of Examples in GSCE}

\begin{figure}[h]
    \centering
    \includegraphics[width=1\linewidth]{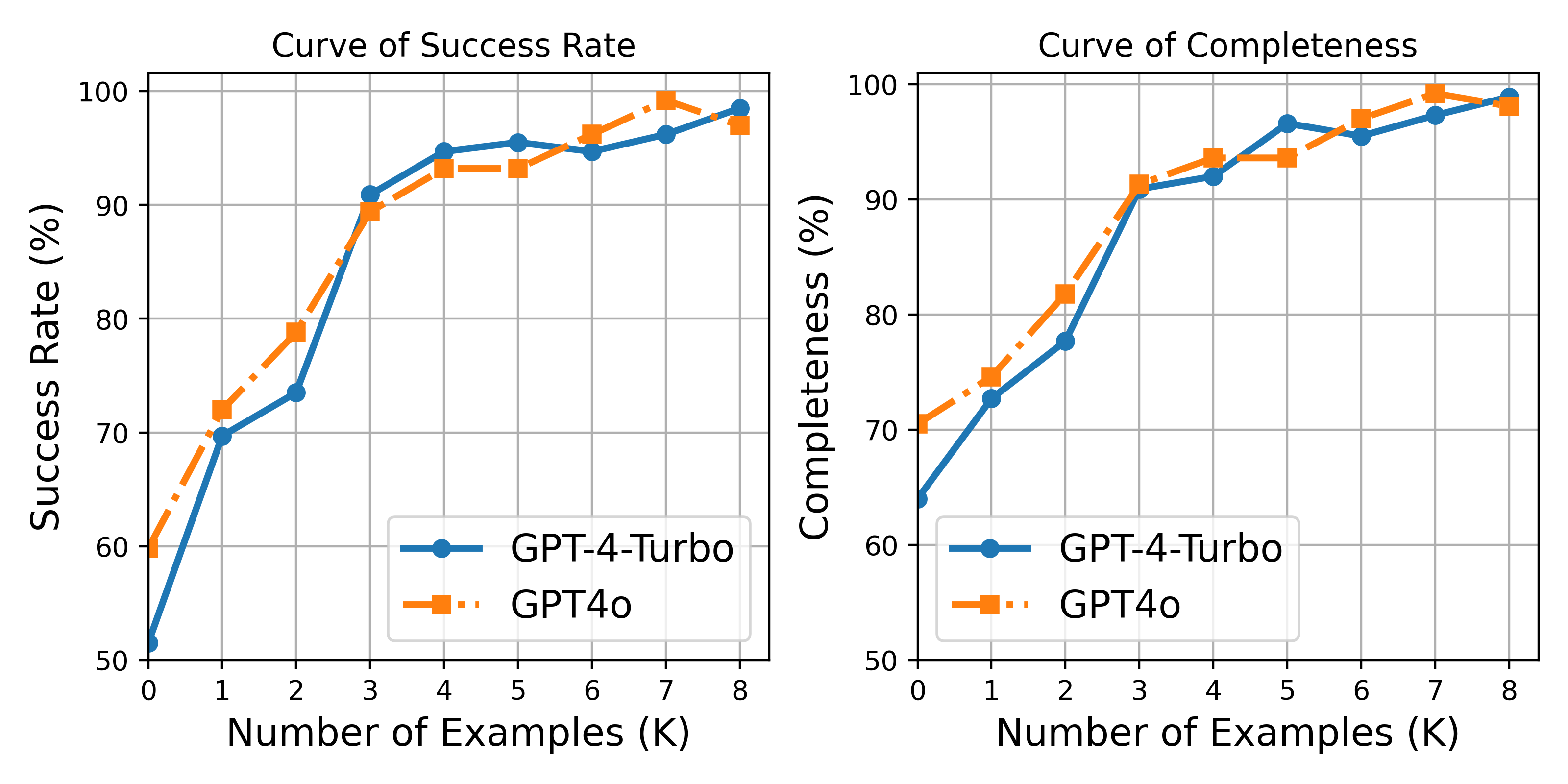}
    \caption{Success Rate and Completeness across varying numbers of examples in the GSCE framework. Both GPT-4-Turbo and GPT-4o exhibit performance improvements as the number of examples increases. However, the improvement becomes marginal when the number of examples exceeds three.}
    \label{fig:SR&Completeness}
\end{figure}

The evaluation result regarding the influence of the number of examples in the GSCE framework is visualized in Figure \ref{fig:SR&Completeness}. The plots indicate that both GPT-4-Turbo and GPT-4o show consistent improvements in success rate and completeness as the number of examples increases, highlighting the effectiveness of in-context learning in LLMs. In both metrics, GPT-4o initially outperforms GPT-4-Turbo at lower example counts (\(K\leq2\)), but GPT-4-Turbo surpasses it after \(K=3\). Interestingly, GPT-4o achieves slightly higher performance at larger example counts (\(K\geq6\)). Both metrics stabilize at high levels (above 90\%) after \(K=4\), with diminishing returns as more examples are added. These findings suggest that while additional examples enhance performance, the gains become marginal when the number of examples exceeds an optimal range. Furthermore, incorporating more examples results in higher OpenAI credit consumption. Consequently, we recommend using three to four examples as the optimal configuration in the GSCE framework to balance performance and token generation cost.

\section{Conclusion \& Future Work}
In this paper, we proposed the GSCE prompt framework to enhance the reasoning capabilities of LLMs for drone control. By integrating guidelines, skill APIs, constraints, and examples, our approach significantly improved success rates and task completeness compared to other methods. Experimental results demonstrated that the combination of NL constraints and examples maximized the reasoning performance of LLMs, with the GSCE framework achieving over 90\% success rate. These findings highlight the critical role of integrating NL constraints with in-context learning examples to enhance reasoning performance. While prior methods relied solely on either NL constraints or examples, our results demonstrate that combining both leads to superior reasoning capabilities.

While our framework substantially enhances LLM-driven drone control, several challenges remain. Future work will explore the following directions:

\begin{itemize}
    \item Integration with Multimodal Inputs: Future work will explore the integration of visual and spatial reasoning capabilities by incorporating multimodal inputs (e.g., images, LiDAR) to enable LLMs to make semantic-aware control decisions.
    \item Adaptive Constraint Learning and Generalization: While predefined constraints improve reasoning reliability, incorporating adaptive constraints that evolve based on dynamic and unpredictable conditions could further enhance safety and performance.
\end{itemize}

By addressing these challenges, we aim to further advance LLM-powered autonomous drone control, enabling reliable drone systems.

\section*{Acknowledgment}
This work is supported by the US National Science Foundation awards 2318710 and 2318711.

\bibliographystyle{unsrt}
\bibliography{ref}

\end{document}